\documentclass[letterpaper, 10 pt, conference]{ieeeconf}
\IEEEoverridecommandlockouts 
\usepackage[utf8]{inputenc}
\usepackage[misc]{ifsym}
\usepackage{color}
\usepackage{enumerate}
\usepackage{tabularx} 
\usepackage{arydshln}
\usepackage{graphics} 
\usepackage{mathptmx} 
\usepackage{times} 
\usepackage{multirow}
\usepackage{amsmath} 
\usepackage{amssymb}  
\usepackage{graphicx}
\usepackage{algorithm, algorithmic}
\usepackage{subcaption} 
\usepackage{helvet}         
\usepackage{courier}        
\usepackage{type1cm}        
\usepackage{makeidx}         
\usepackage{bm}
\usepackage{comment}
\usepackage{notoccite}
\usepackage[font=small,labelfont=bf,tableposition=top,hypcap=false]{caption}

\usepackage{ulem}

\newcommand{\etal}{\textit{et al.~}}
\def\eg{\textit{e.g.},~}

\makeatletter
 \let\NAT@parse\undefined
 \makeatother
\usepackage[pdftex,
colorlinks=true,
bookmarks=true,
citecolor=red,
linkcolor=blue,
pagebackref=true,
 pdfstartview=Fit,
  breaklinks=true
]{hyperref}

\title{
FlowFusion: Dynamic Dense RGB-D SLAM Based on Optical Flow}
\author{
    Tianwei Zhang$^{1}$,
     Huayan Zhang$^{2}$, 
     Yang Li$^{1}$,
    Yoshihiko Nakamura$^{1}$
    and Lei Zhang$^{1,2}$
    \thanks{
        $^{1}$ Department of Mechano-Informatics, School of Information Science and Technology, the University of Tokyo, 7-3-1 Hongo, Bunkyo-ku, Tokyo, Japan.    
        \{zhang, nakamura\}@ynl.t.u-tokyo.ac.jp, liyang@mi.t.u-tokyo.ac.jp}
    \thanks{ $^{2}$ School of Electrical and Information Engineering, Beijing University of Civil Engineering and Architecture, No.1 Zhanlanguan Road, Xicheng District, Beijing, China.
        \{hyzhang, leizhang\}@bucea.edu.cn}
}

\begin{document}
\maketitle
\begin{abstract}
    Dynamic environments are challenging for visual SLAM since the moving objects occlude the static environment features and lead to wrong camera motion estimation. 
    In this paper, we present a novel dense RGB-D SLAM solution that simultaneously accomplishes the dynamic/static segmentation and camera ego-motion estimation as well as the static background reconstructions.
    Our novelty is using optical flow residuals to highlight the dynamic semantics in the RGB-D point clouds and provide more accurate and efficient dynamic/static segmentation for camera tracking and background reconstruction. 
   The dense reconstruction results on public datasets and real dynamic scenes indicate that the proposed approach achieved accurate and efficient performances in both dynamic and static environments compared to state-of-the-art approaches.
\end{abstract}

\section{Introduction}
\label{1}
Simultaneous Localization and Mapping (SLAM) method for a robot is to acquire the information from the unknown environment, build up the map and locate the robot itself on that map. 
Dynamic environment is a big problem for the real scene implementation of SLAM in both robotics and computer vision research fields.  
The reason is that most of the existed SLAM approaches and Visual Odometry(VO) solutions guarantee their robustness and efficiencies based on the static environment assumption. 
When the dynamic obstacles occur or the observed environment changes, these methods cannot extract enough reliable static visual features, so as to insufficient feature associations, which lead to the motion estimation failures between different camera poses.

To deal with the dynamic environments, one straightforward idea for visual SLAM is to extract the dynamic components from the input data and filter them as exceptions to apply the existed robust static SLAM frameworks. 
Recently, the fast development of deep learning-based image segmentation and object detection methods have gained greatly in both efficiency and accuracy. Many researchers try to handle the dynamic environments via involving semantic labeling or object detection pre-processing to remove the potential dynamic objects. These methods have shown very effective results in particular scenes dealing with particular dynamic objects. 
However, their robustness may drop down when unknown dynamic objects turn up. Considering more generalized dynamic features, flow approaches are explored to describe all kinds of dynamic objects, \eg the scene flow in 3D point clouds and the optical flow in 2D images. The flow approaches are to estimate the pixel motions between the given image pair or point clouds data. These methods are sensitive to slight motions and have advantages when tracking the moving non-rigid surfaces. Nevertheless, flow methods need complex penalty setting and suffered from the unclear segmentation boundaries. 

In this paper, to get rid of the pre-known dynamic object hypothesis, we deal with the dynamic SLAM problem via flow based dynamic/static segmentation. Different from the existed methods, we provide an novel optical flow residuals based dynamic segmentation and dense fusion RGB-D SLAM scheme. 
Through improving the dynamic factor influence, in our approach, the dynamic segments are efficiently extracted in current RGB-D frame, the static environments are then accurately reconstructed. Moreover, demonstrations on the real challenging humanoid robot SLAM scenes indicate that the proposed approach outperforms the other state-of-the-art dynamic SLAM solutions.

\section{Related Works}
\label{2}
\begin{figure*}[tbh]
    \centering
        \includegraphics[width=1.8\columnwidth]{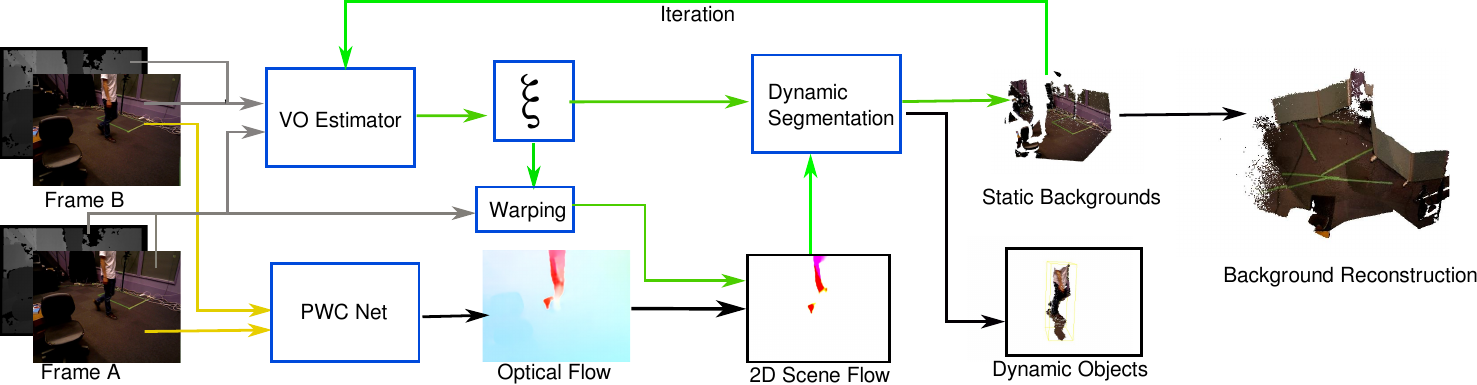}
        \caption{The proposed FlowFusion framework:
       Input two continuous RGB-D frames A and B, the RGB images are first fed into PWC-net for optical flow (yellow arrows) estimation. Meanwhile, the intensity and depth pairs of A and B are fed to robust camera ego-motion estimator to initialize the camera motion $\xi$ (introduced in Section \ref{sec:robustEgo}). We then warp the frame A to A' with $\xi$ and obtain the projected 2D scene flow (Section \ref{sec:of}), then apply it to dynamic segmentation. After several iterations(Section \ref{sec:seg}, the green arrows), the static backgrounds are achieved for reconstruction. 
     }
\label{chart}
\vspace{-0.5cm}
\end{figure*}

Saputra \etal summary the dynamic SLAM methods by the year of 2017 in \cite{dynamicSLAM-survey}. 
Most of these approaches are dedicated to specialized scenes.
For human living environments, benefit from the economic RGB-D sensors and the computational power improvement from economical graphics processing units(GPU), 
the dense RGB-D fusion based approaches, \eg KinectFusion \cite{kf} and ElasticFusion(EF) \cite{ef}, made real-time static indoor environments reconstruction come true with high robustness and accuracy. 
Many researchers tried to extend these frameworks to dynamic scenes:

The motion segmentation problem can be treated as a semantic labeling problem. For instance, R. Martin \etal proposed Co-Fusion(CF) in \cite{cf} and Xu \etal proposed Mid-Fusion in \cite{mid-f}.
CF is a real-time object segmentation and tracking method which combined the hierarchical deep learning based segmentation method from \cite{cf-seg} and the static dense reconstruction framework of EF.

Besides the semantic labeling solutions, some people insisted to find out the dynamic point clouds as outliers from the dense RGB-D fusion scheme. Such as J, Mariano \etal provided a joint motion segmentation and scene flow estimation method(JF) in \cite{jf}, R. Scona \etal proposed a static backgrounds reconstruction approach in StaticFusion(SF) \cite{sf}.

In addition, with the help of deep learning based object detection approaches, some works deal with the dynamic environment problem by involving object detection pre-processing and remove the potential dynamic objects, then reconstruct the environments via static SLAM frameworks. \eg Zhang \etal proposed the human object detection and background reconstruction method PoseFusion(PF) \cite{pf}; C. Yu \etal in \cite{ds} applied SegNet \cite{segnet} to detect and remove foreground humans and then estimate the camera motions with ORB-SLAM2 \cite{orb2} framework.

More than that, some researchers tried to define the environment dynamic properties as a semantic concept
and solve it with SLAM tools.
The environment rigidity is firstly defined as a semantics instead of the particular object classification in \cite{rigidity}. In which, the environment rigidity refers to the static background point clouds set, which is stationary, as opposed to the moving objects. 
Then, in \cite{lvzhaoyang}, Lv \etal proposed a deep learning based 3D scene flow estimation approach, which combines two deep learning networks: the optical flow approach from \cite{pwc}, and another net for static background rigidity learning.
       

\section{Optical Flow based Joint Dynamic Segmentation and Dense Fusion}
This Section describes how does the proposed VO keep its robustness in dynamic environments. The flowchart of the proposed approach is shown in Fig.\ref{chart}. Our approach takes two RGB-D Frames A and B as input, the RGB images are fed into PWC to estimate the optical flow(yellow arrows). Meanwhile, the intensity and depth pairs of A and B are fed to the robust camera ego-motion estimator to estimate an initial camera motion $\xi$ (introduced in Section \ref{sec:robustEgo}). We then warp frame A to A' with $\xi$ and obtain the projected 2D scene flow(in Section \ref{sec:of}) for dynamic segmentation. After several iterations(Section \ref{sec:seg}, the green arrows), the static backgrounds are achieved for the following environment reconstruction. As the proposed method applied optical flow residuals for dynamic segmentation, we name it as FlowFusion (FF).

\subsection{Visual Odometry in Dense RGB-D Fusion}
\label{sec:robustEgo}
Following the RGB-D fusion frameworks \cite{ef} and \cite{jf}, our VO front-end is formulated as an optimizing problem of the color(photometric) and depth(geometric) alignment errors. 
Given RGB-D camera frames $A$ and $B$,
If we denote intensity image as $I$, depth image as $D$, $I_{A} \in \mathbb{R}^2$ and $D_{A} \in \mathbb{R}^2$ are the intensity and depth images on the 2D image plane of camera frame $A$. 
The 3D Point Clouds Data(PCD) can be generated from ($I_A, D_A$) via pinhole camera model.
We first over segment the PCD of $A$ into $N$ 
clusters $\mathcal{V} =\Sigma_{i=1}^N V_i $
according to supervoxel clustering \cite{sv} and obtain the adjacency graph $\mathcal{G}\{\mathcal{V}, E_{ij}\}$ (similar to JF and SF, for efficiency, we use intensity distance instead of RGB distance for clustering).
As each cluster $V_i$ is composed of similar point clouds, we treat each cluster as a rigid body. 
Then, we define $\xi \in \mathfrak{se}(3)$ as the initial rigid motion guess of the that frame. 
Assume that the robot starts to move in a static environment, $\xi$ can be solved from the formulated energy function considering photometric and depth residuals:
\begin{equation}
\xi = {arg~min}_{\xi} \{ \sum_{p=1}^{N}[C(\alpha_Iw_I^p r^p_I(\xi)) + C(w_D^pr^p_D(\xi))]
\}
\label{eq:minimize} 
\end{equation} 
in which, $\alpha_I $ is a scale factor to make the intensity item be comparable to the depth term. The $w_D$ and $w_I$ pre-weight the depth and intensity terms according to their measurement noise. 
The photometric residuals $r_I$ can be computed as:
\begin{equation}
r^p_I(\mathbf{\xi}) = I_B(\mathcal{W}(\mathbf{x}^p,\mathbf{\xi})) - I_A(\mathbf{x}^p)
\end{equation}  
and the geometric residuals $r_D$ are obtained from the Depth measurements. 
\begin{equation}
r^p_D(\xi) = D_B(\mathcal{W}(\mathbf{x}^p,\xi))- |T(\xi)\pi^{-1}(\mathbf{x}^p, D_A(\mathbf{x}^p))|_D
\end{equation} 
the transformation is denoted as $T(\xi) \in {SE}(3)$, which is the transformation of $\xi \in \mathfrak{se}(3)$, it is composed
by the camera rotation and translation between the $A$ and $B$ frames. In addition, the $\mathcal{W}$ stands for an image warping operation:
\begin{equation}
\mathcal{W}(\mathbf{x},\xi) = \pi(T(\xi)\pi^{-1}(\mathbf{x}, D_A(\mathbf{x})))
\end{equation}
in which, the $\mathbf{x}^p$ stands for the pixel coordinates on the 2D image of $p$. $|\cdot|_D$ indicates the depth value on the depth image.
We denote $\pi$ as the projecting from a world coordinate point to camera plane, and we denote the extrinsic parameters as $T(\xi)$.
\begin{equation}
\pi:\mathbb{R}^3\rightarrow \mathbb{R}^2 : 
\end{equation}
the function of $\pi$ is depending on the sensors types (such as pinhole camera, stereo camera, and laser scanners). In this paper, we deal with RGB-D PCD, thus it is a pinhole camera model here. 
\begin{figure}
    \centering
        \includegraphics[width=0.75\columnwidth]{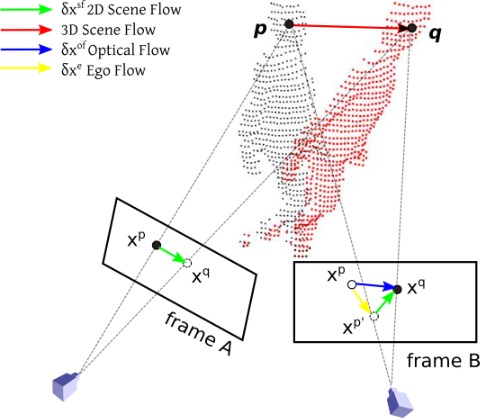}
        \caption{The projected 2D scene flow in image planes. $x^p$ is an object point project pixel in frame A and $x^q$ is the same 3D point (belong to the moving object) on frame B. The red arrow indicates the scene flow, which is the world space motion, the blue arrows are the optical flows $x_{A->B}^{of}$, the green arrows are the projected 2D scene flows $x^{sf}$ on image planes, the yellow vectors are the ego flows $x^e$ resulted from camera ego-motions.
}
\label{fig:of}
\vspace{-0.5cm}
\end{figure}

Finally, the function of $C(r)$ is a robust penalty to balance the optimization computation's robustness and convergence. 
For RGB-D visual odometry, refers to \cite{jf, sf}, the Cauchy robust penalty is usually adopted since it's more robust than $L_1/L_2$ norms:
\begin{equation}
C(r) =\frac{c^2}{2} log(1+(\frac{r}{c})^2)
\end{equation} 
in which, the $c$ is the inflection point of $F(r)$, which can be tuned according to the residual levels. Equation~\ref{eq:minimize} is high nonlinear, we solve it via coarse-to-fine scheme using the iterative re-weighted least-square solver provided by \cite{robust}. 
This VO estimator works well in static environments but loses its robustness in dynamic cases. The reason is that, in Eq.\ref{eq:minimize}, the depth and intensity residuals contribute to the VO estimator based on the environment's rigid motion hypothesis. To deal with the dynamic objects, we define the optical flow residuals which directly indicates the non-rigid environment motions. 

\subsection{Optical Flow Residual Estimated by Projecting the Scene Flow}
\label{sec:of}
\begin{figure*}
   \begin{subfigure}{0.5\columnwidth}
      \centerline{\includegraphics[width=0.8\columnwidth]{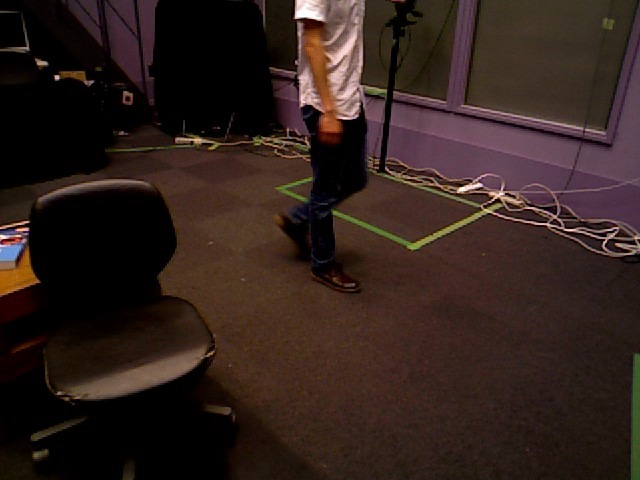}}
       \centering {(a) Dynamic Scene}\medskip
    \end{subfigure}
   \begin{subfigure}{0.5\columnwidth}
      \centerline{\includegraphics[width=0.8\columnwidth]{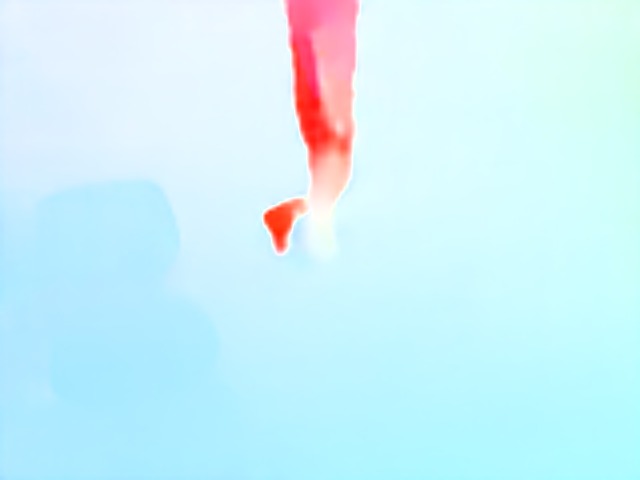}}
    \centering {(b) Optical flow}\medskip
    \end{subfigure}
   \begin{subfigure}{0.5\columnwidth}
      \centerline{\includegraphics[width=0.8\columnwidth]{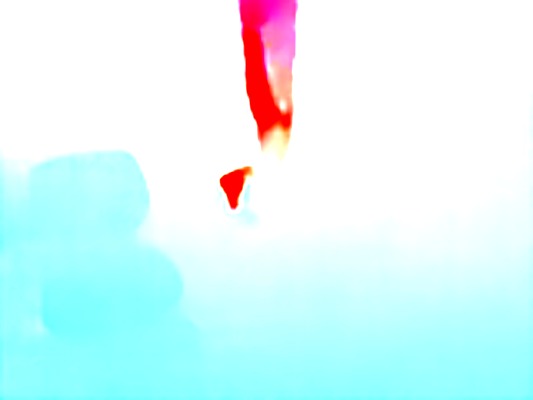}}
       \centering {(c) 2D scene flow}\medskip
    \end{subfigure}
   \begin{subfigure}{0.5\columnwidth}
      \centerline{\includegraphics[width=0.8\columnwidth]{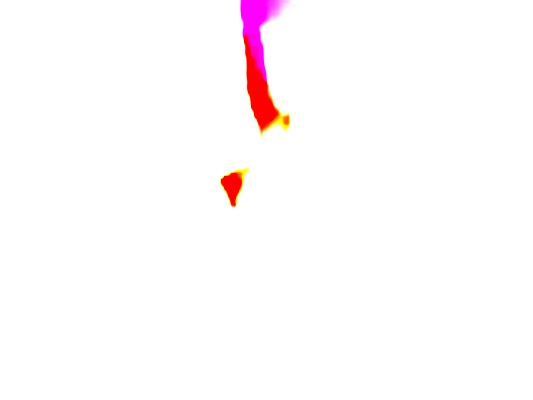}}
    \centering {(d) Iteration 7}\medskip
    \end{subfigure}
     \caption{Iteratively estimate the 2D scene flows in a dynamic scene. (a) is the scene, in which the robot was moving leftwards and the human was moving rightwards. (b) is the optical flow estimated from the image pair of (a). The colors indicate flow direction, the intensity indicates the pixel displacement. The blue flows resulted from camera ego-motion. 
We subtract the ego flows from the optical flows, and obtain the scene flow components on the image plane as shown in (c). Iteratively remove the scene flows and ego flows in (b), the better 2D scene flow results can be achieved as (d) after 7 iterations.}
    \label{fig:2dflow}
    \vspace{-0.5cm}
 \end{figure*} 

Theoretically, we can distinguish a cluster $V_i$ is dynamic or static via $\xi$.
As ($I_A, D_A$) are warped using $\xi$ then we compute the average residuals of each cluster, the real backgrounds do not move, their pixel clusters move along with the camera motion $\xi$, thus their residuals are low. The dynamic clusters which move along with the dynamic objects should contain high residuals since their motions don't coincide with the camera motion $\xi$.
Therefore, the dynamic clusters can be extracted by setting the thresholds for high and low $r_I, r_D$ residuals. 
However, in the real cases, the intensity and depth residuals are not good metrics, the reasons are:
\begin{itemize}
\item The depth and intensity images are obtained from different lens, they cannot be registered perfectly since the time delay.
\item The depth measurement is discrete on the boundary regions, which results in wrong alignment.
\item The depth measurement errors grow along with the range.
\end{itemize} 
 
To deal with these problems, we want to find a concept that directly indicate the pixel or point clouds' dynamic level. The Scene flow method is to estimate the moving 3D points, but it cannot be  obtained directly (\eg in JF, the scene flows were obtained after several VO estimation iterations). On the other hand, optical flows which can be easily obtained from image pairs are often applied to describe the moving objects captured by static cameras. 
Therefore, to get rid of the camera ego-motions, inspired by \cite{lvzhaoyang}, 
we involve the concept of optical flow residual, which is defined as projected 2D scene flow, to highlight the pixel's dynamic property.

Specifically, to estimate the optical flow between time $t$ and $t+1$:

\begin{equation}
\begin{split}
\delta \mathbf { x } _ { t \rightarrow t + 1 } ^ { o f } =~  
& \pi \left( T _ { t + 1 } (\mathbf { x } _ { t } + \delta \mathbf { x } _ { t \rightarrow t + 1 }, D\left( \mathbf { x } _ { t } + \delta \mathbf { x } _ { t \rightarrow t + 1 } \right)\right) \\ 
&- \pi \left( T _ { t } (\mathbf {x}_{ t }, D(x_t)) \right)
\end{split}
\label{eq:of}
\end{equation}

See Figure \ref{fig:of}, $x^p$ is an pixel of an object point in frame A and $x^q$ is the same object point seen in frame B. 
The red arrow indicates the scene flow, which is a 3D motion in world space. The blue arrows are the optical flows $x^{of}$, the green arrows are the projected 2D scene flows $x^{sf}$ on image planes, the yellow vectors are the camera ego-motion flows $x^e$.

The optical flows are defined as the pixel motions on the image coordinates, as shown in Figure \ref{fig:2dflow} (b), in which, the colors indicate flow direction and the intensity indicate the pixel displacement. 
In the real scene of (a), the robot was moving leftwards and the human was moving rightwards. Thus the blue flows were resulted from camera ego-motion. 
We define such kind of flow as the camera ego flow $\delta \textbf{x}^e$, which means the observed optical flow was purely resulted from camera motion (without moving objects). If we subtract the ego flows from the optical flows, the scene flow components on the image plane can be obtained, as shown in (c) and (d).

For one 2D pixel $x$ of frame $A$, given the camera motion $\xi \in \mathfrak{se}(3)$, the camera ego-motion flow can be computed as:
\begin{equation}
\delta \textbf{x}^e_{ A \rightarrow B } = \mathcal{W}(\mathbf{x}, \xi)
- \mathbf { x } 
\end{equation}

The projected scene flow on the image plane can be computed as:
\begin{equation}
\delta \mathbf {x} _ {A\rightarrow B} ^ {sf} = \delta \mathbf {x} _ { A \rightarrow B } ^ { o f } - \delta \mathbf { x } _ { A \rightarrow B } ^ {e}
\label{sceneflow}
\end{equation}

For the static pixels, Equation \ref{sceneflow} is close to zero, since its optical flow comes from the camera motions. For the dynamic pixels, the 2D scene flows are non-zero, and their absolute values grow along with the moving speed.
Therefore, we define the flow residual $r_F(x^p)$ as its corresponding $\delta \mathbf {x} _ {A\rightarrow B} ^ {sf}$.
As the dense optical flow computation is time-consuming, instead of using Equation \ref{eq:of}, we apply a GPU speed-up dense optical flow estimation method PWC-net \cite{pwc}.

\subsection{Dynamic Clusters Segmentation}
\label{sec:seg}
By now, we have projected the frame $A$ using the VO $\xi$, we have defined three residuals, $r_I, r_D$ and $ r_F$, relative to the intensity, depth and optical flow, respectively.
We proposed to distinguish a cluster is static or not according to its average residuals. This will be done in two procedures.
Firstly, we compute a metric to combine these three residuals, secondly, we compose a minimizing function to qualify the dynamic level of clusters.

To combine the residuals, an average residual $\delta_i$ of the cluster $i$ is defined as:
\begin{equation}
\delta_i = \Sigma_{n=1}^{S_i}(\alpha_Ir_I^n + r_D^n/D_i + \alpha_Fr_F^n) 
\end{equation}
in which $S_i$ is cluster size, $D_i$ is the cluster's average depth, $\alpha_F$ and $\alpha_I$ control the flow and intensity weights.
For each cluster, we compute its dynamic level $b_i \in [0,1]$
$b_i = 0$ means cluster $i$ is definitely belong to static segments.

Then we formulate the energy function of clusters $\textbf{b}$:
\begin{equation}
E(\textbf{b}) = E_{\delta}(\textbf{b}) + E_G(\textbf{b})
\label{E(b)}
\end{equation}
in which, $E_{\delta}(\textbf{b})$ stands for the relationship between $b_i$ and the threshold. Let's set top and bottom average residuals as $\theta_t, \theta_b$, then,
\begin{equation}
E_{\delta}(\textbf{b}) = \Sigma_{n=1}^{N}w(\delta_i)(b_i - g(\delta_i))^2
\end{equation}
respect to the assignment function:
\begin{equation}
g \left( \delta _ { i } \right) = \left\{ \begin{array} { l l } { 0 } & { \delta _ { i } < \theta _ { b } } \\ { \left( \delta _ { i } - \theta _ { b } \right) / \left( \theta _ { t } - \theta _ { b } \right) } & { \theta_ { b } \leq \delta _ { i } \leq \theta _ { t } } \\ { 1 } & { \delta _ { i } > \theta _ { t } } \end{array} \right.
\end{equation}

To increase the contribution of high residual parts, the weight $w_\delta$ is defined as:
\begin{equation}
w \left( \delta _ { i } \right) = \sqrt { \left( \frac {  \delta _ { i } -  \theta _ { b }  } { \theta _ { t } - \theta _ { b } } \right) ^ { 2 } + 1 }
\end{equation}

Refer to SF~\cite{sf}, to enhance the connectivity of adjacent clusters and push the similar clusters to the same dynamic/static segments, we form $E_G$:
\begin{equation}
E _ { G } ( \boldsymbol { b } ) = \sum _ { i = 1 } ^ { N } \sum _ { j = i + 1 } ^ { N } G _ { i j } \left( b _ { i } - b _ { j } \right) ^ { 2 }
\end{equation}
with the supervoxel adjacency graph: $\mathcal{G}\{\mathcal{V}, E_{ij}\}$.
$G_{ij} = 0$ if $E_{ij} = 0$, otherwise  $G_{ij} = 1$.

As $E_G(b)$ is convex, since it is designed with all squared items. Thus the Equation \ref{E(b)} could be solved respect to $\textbf{b}$.
Once obtaining $\textbf{b}$, we then modify the Equation \ref{eq:minimize} to considering the dynamic-static segmentation:
\begin{equation}
\xi = {arg~min}_{\xi} 
\{\sum_{p=1}^M(1-b_i(p))[C(\alpha_Iw_I^p r^p_I(\xi)) + C(w_D^pr^p_D(\xi))]
\}
\label{eq:final}
\end{equation}
in which, $b_i(p)$ is the dynamic score of cluster $i$ which contains the pixel $x^p$. $M$ is the size of static pixels. 
We can solve this Equation \ref{eq:final} with the solved $\textbf{b}$ using the iteratively re-weighted least-square solver provided by \cite{dual, robust}.
\begin{figure*}[t]
    \centering
        \includegraphics[width=1.5\columnwidth]{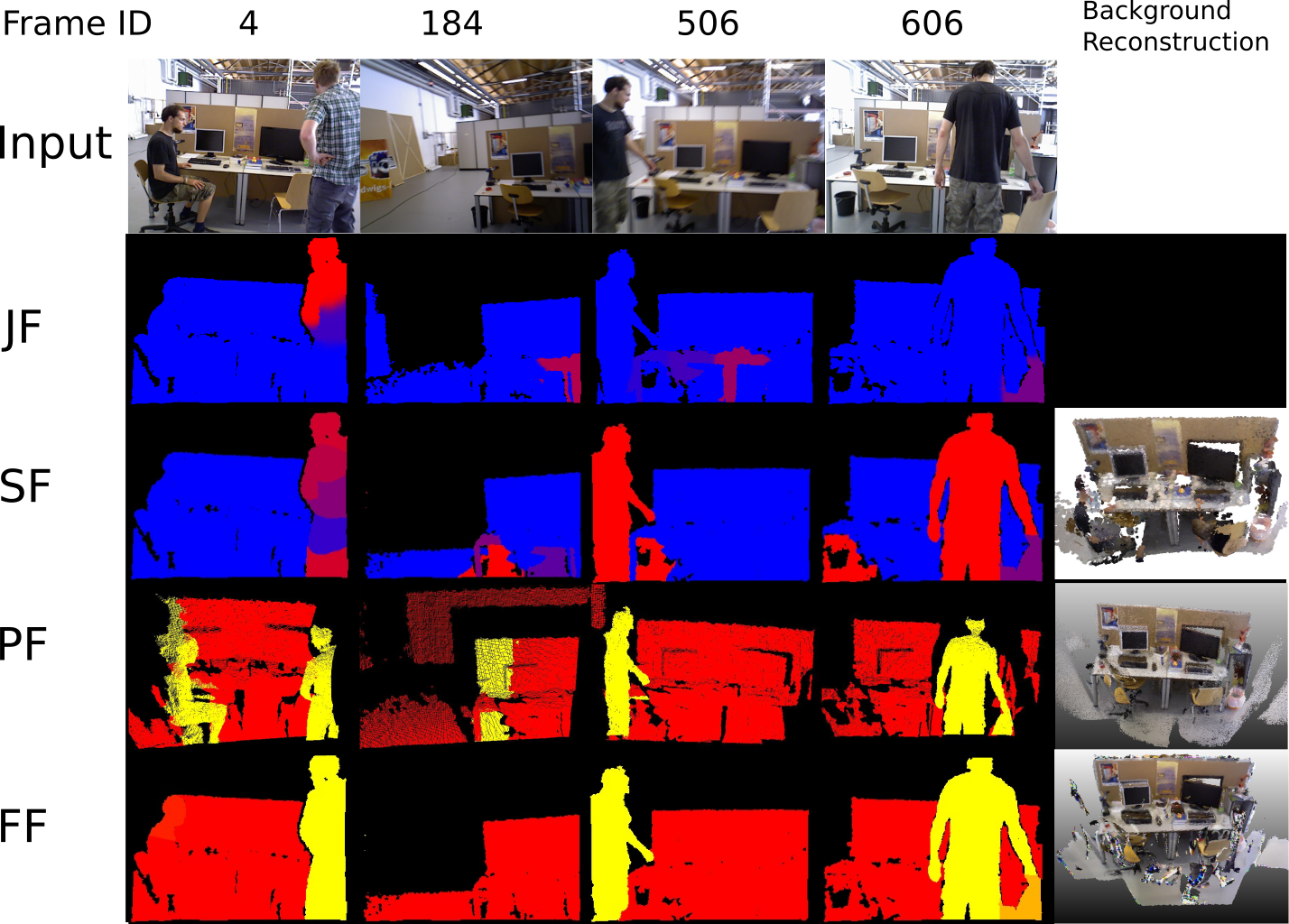}
        \caption{Comparison Experiments on TUM $fr3/walking\_xyz$ sequence. The dynamic segmentation performances of JF, SF, PF and the proposed FF approaches are compared. The blue parts are static in JF and SF. The red Parts are static in PF and FF. The first row is input RGB frames, the other rows are the dynamic/static segmentation results of each method and the last column show the background reconstructions(except JF, which didn't provide reconstruction function). See Tab.\ref{tab1} and Tab.\ref{tab2} for the comparison results. 
     }
\label{fig:tumseg}
\vspace{-0.5cm}
\end{figure*}

\begin{figure*}[tb]
	\begin{subfigure}{0.5\columnwidth}
    \centering
       \includegraphics[width=\columnwidth]{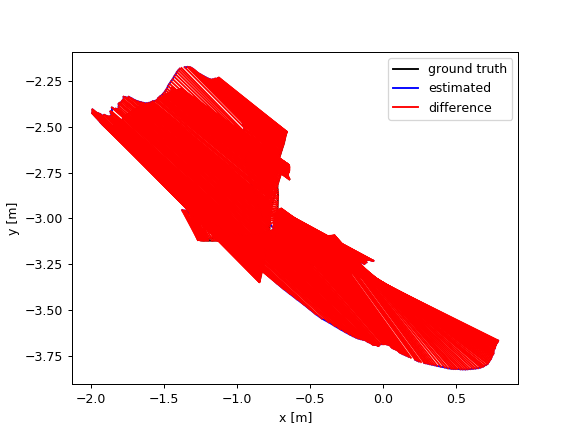}
        \caption{JF ATE}
   		\end{subfigure}%
\hfill
 	\begin{subfigure}{0.5\columnwidth}
    \centering
       \includegraphics[width=0.9\columnwidth]{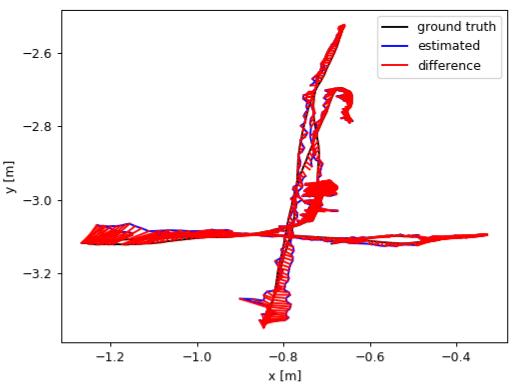}
        \caption{PF ATE}
                \vspace{-0.3cm}
 		\end{subfigure}%
 			\begin{subfigure}{0.5\columnwidth}
    \centering
       \includegraphics[width=\columnwidth]{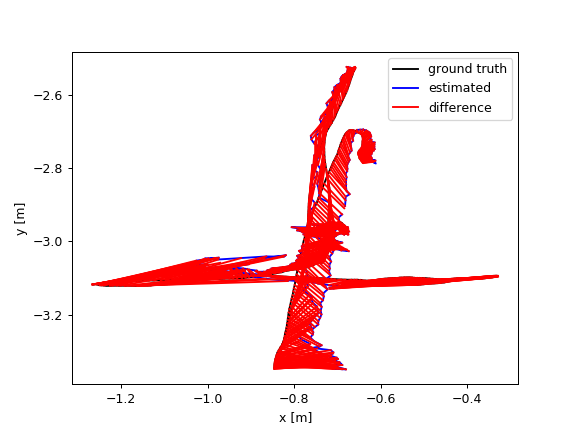}
        \caption{SF ATE}
   		\end{subfigure}%
\hfill
 	\begin{subfigure}{0.5\columnwidth}
    \centering
       \includegraphics[width=\columnwidth]{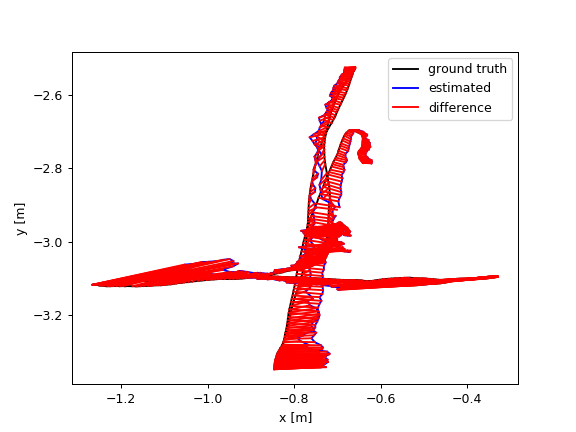}
        \caption{FF ATE}
 		\end{subfigure}%
 \hfill 
 	\begin{subfigure}{0.5\columnwidth}
    \centering
       \includegraphics[width=\columnwidth]{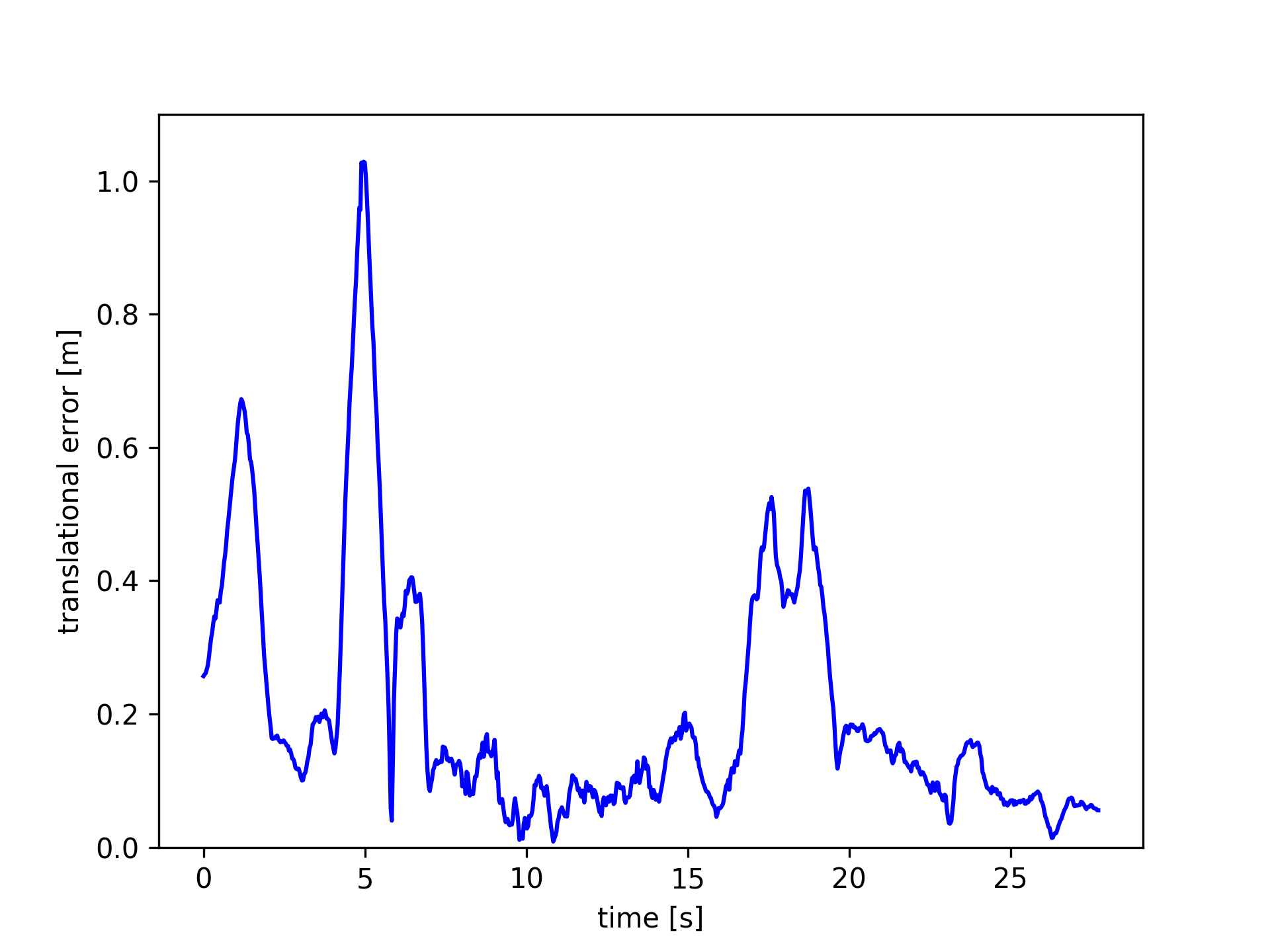}
        \caption{JF RPE}
   		\end{subfigure}%
\hfill
 	\begin{subfigure}{0.5\columnwidth}
    \centering
       \includegraphics[width=\columnwidth]{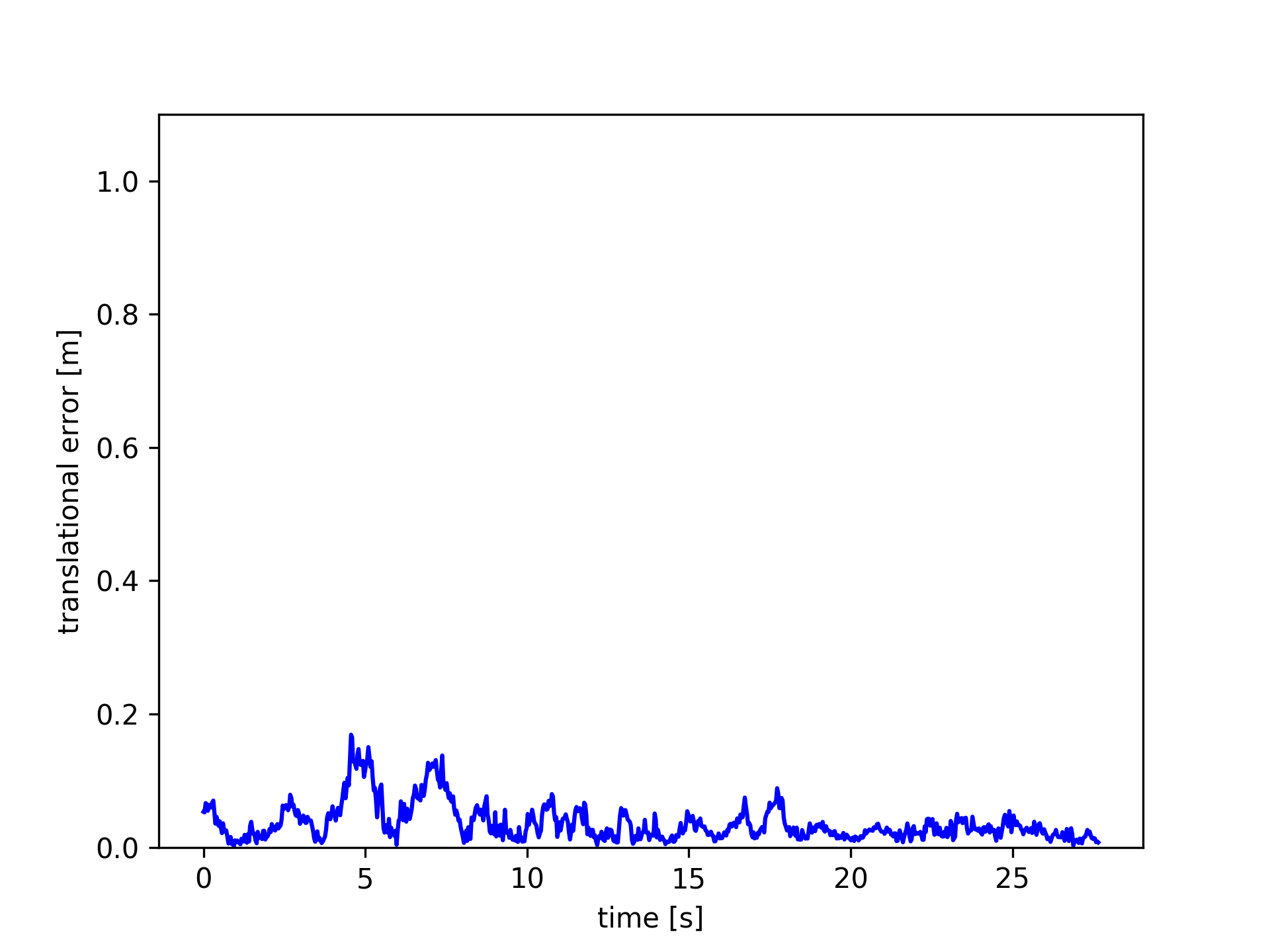}
        \caption{PF RPE}
 		\end{subfigure}%
 			\begin{subfigure}{0.5\columnwidth}
    \centering
       \includegraphics[width=\columnwidth]{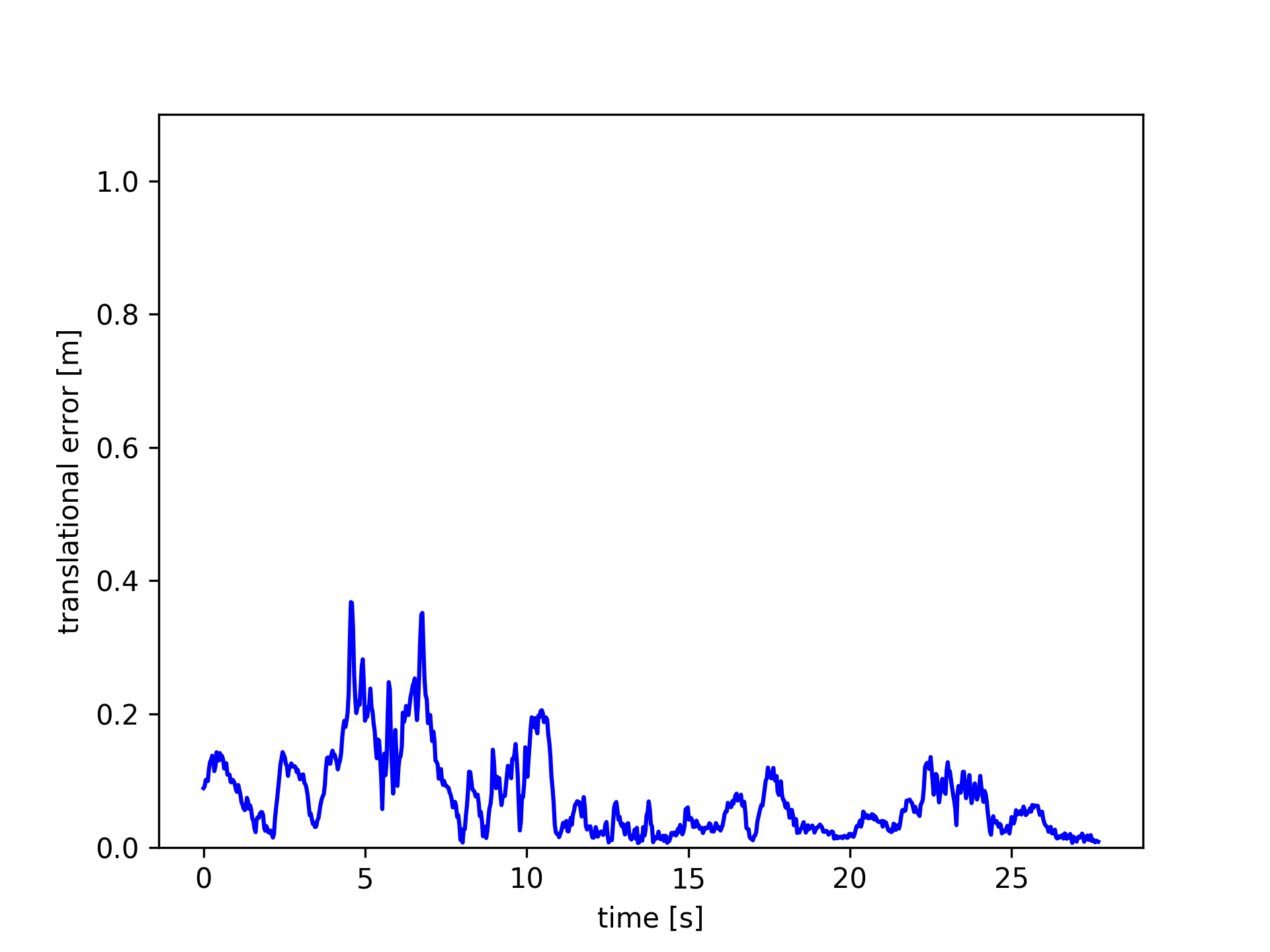}
        \caption{SF RPE}
   		\end{subfigure}%
\hfill
 	\begin{subfigure}{0.5\columnwidth}
    \centering
       \includegraphics[width=\columnwidth]{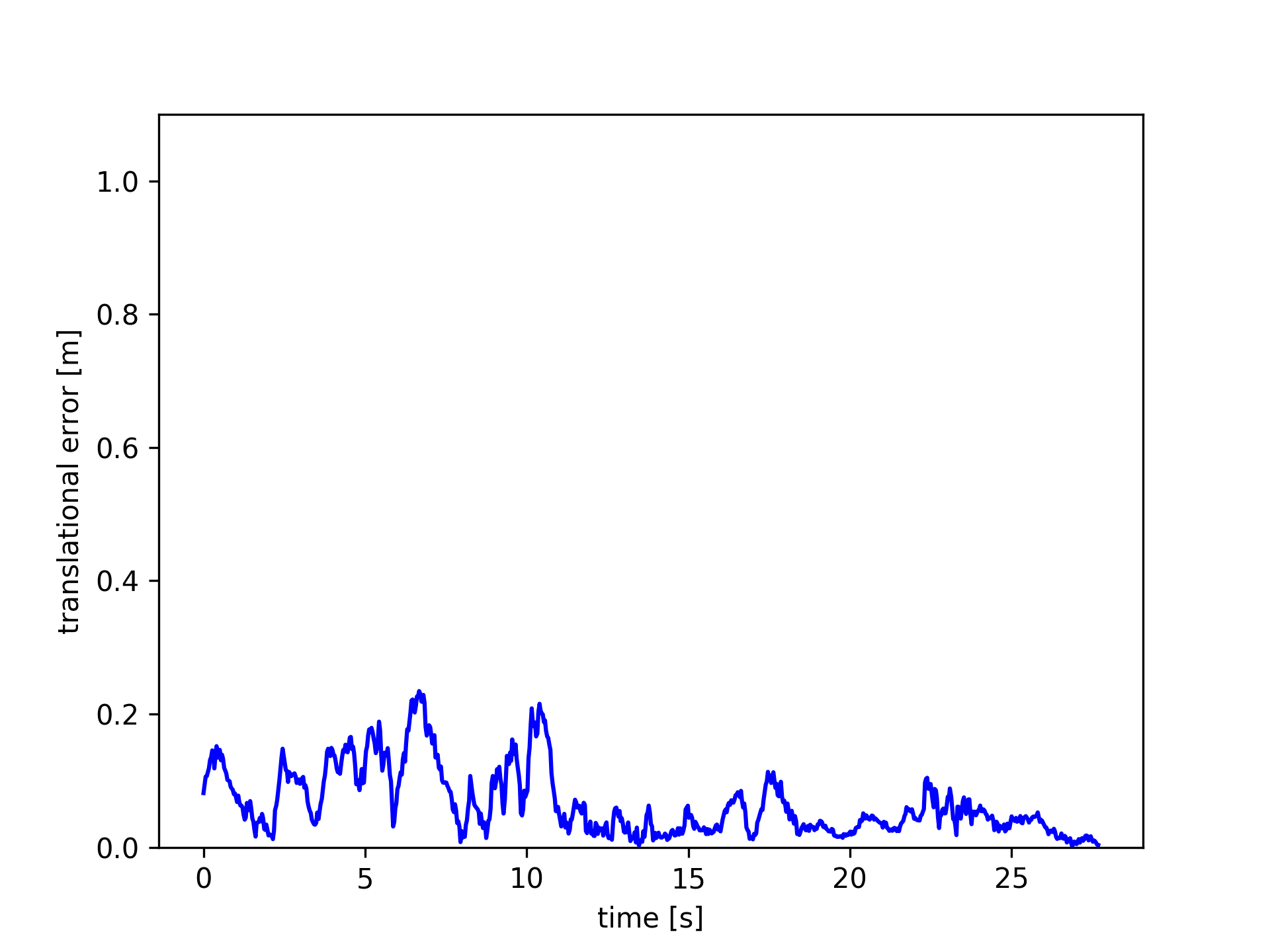}
        \caption{FF RPE}
 		\end{subfigure}%
\caption{ATE and RPE of TUM fr3/walking\_xyz dynamic sequence.  The object detection based PF achieved the smallest trajectory errors.
The proposed FF performs better than the other model-free dynamic SLAM solutions. 
About the ATE RMSE, PF achieves 4.1 cm while FF gets 12 cm.
For the RPE RMSE, PF achieves 13 cm/s while FF get 21 cm/s.}
 	\label{fig:tumplot} 
 	\vspace{-0.5cm}
\end{figure*}
\begin{table}
  \caption{Experimental Parameters List}
  \centering
  \begin{tabular}{|l|c|c|}
      \hline
    TUM & $\alpha_F$ & 0.022\\
      \hline
      TUM &$\alpha_I$ & 0.9 \\
      \hline
       TUM &Max Iteration & 8 \\
         \hline
   HRPSlam & $\alpha_F$ & 0.018\\
      \hline
     HRPSlam &$\alpha_I$ & 0.88 \\
      \hline
    HRPSlam &Max Iteration & 8 \\
      \hline 
  \end{tabular}
  \label{tab:para-list}  
  \vspace{-0.5cm}
\end{table}

\section{Dynamic SLAM Experiments and Evaluations}
\label{sec:exp}
\begin{figure*}[tb]
    \centering
        \includegraphics[width=1.9\columnwidth]{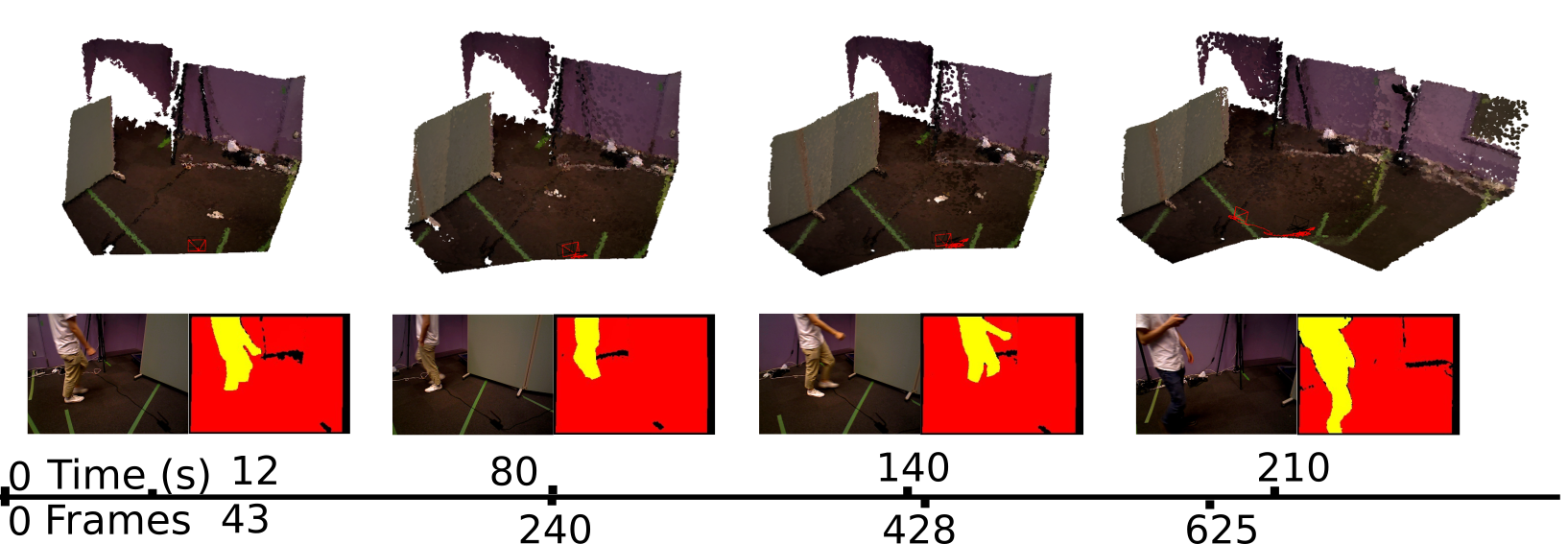}
        \caption{FlowFusion Experimental Result on HRPSlam 2.1 sequence. The yellow parts are the estimated dynamic objects. In this sequence, an HRP-4 humanoid robot mount one RGB-D sensor firstly moved to his left and then turned rightwards. These datasets contaion abundant fast rotation motions and shaking, which make difficulties to obtain optical flow residual. The feet parts are segmented to the static background, since during the walking phase, the supporting feet on the ground are easily treated as static objects. Although the sweeping feet are moving fast and hold significant optical flow residuals, they are too close to the rigid grounds. Thus they are easily segmented to the the static background due to the graph connectivity.}
\label{fig:hrpslam}
\vspace{-0.5cm}
\end{figure*}

To evaluate the proposed FlowFusion dynamic segmentation and dense reconstruction approach, we compare the VO and mapping results of FF to state-of-the-art dynamic SLAM methods SF, JF and PF in the public TUM \cite{tum} and HRPSlam \cite{hrpslam} datasets.
The former provides widely accepted SLAM evaluation metrics: Absolute Trajectory Error (ATE) and Relative Pose Error (RPE).
\begin{table}[b]
\caption{Trans. ATE RMSE (m)}
\label{tab1}  
\begin{tabular}{p{2.4cm}p{1cm}p{1cm}p{1cm}p{1cm}}
\hline\noalign{\smallskip}
\textbf{Sequence}  & JF & SF & FF & PF  \\
\noalign{\smallskip}\hline\noalign{\smallskip}
\hline    
      fr1/xyz & 0.051& \textbf{0.017} &0.020 &0.020\\
      fr1/desk2 & 0.15& 0.051 & 0.034& \textbf{0.023} \\
\hline

 fr3/walk\_xyz& 0.51  & 0.21 & 0.12 & \textbf{0.041}\\
     fr3/walking\_static & 0.35 & 0.037 &\textbf{0.028}& 0.072 \\
     HRPSlam2.1 & 0.51 & 0.25 &0.23 &\textbf{0.21}\\
     HRPSlam2.4 & \textbf{0.32} & 0.44 &0.49 & 0.47\\
     HRPSlam2.6 & 0.21 & 0.18 &\textbf{0.11} &0.15\\ 
\noalign{\smallskip}\hline\noalign{\smallskip}
\end{tabular}
\vspace{-0.5cm}
\end{table}
\begin{table}[b]
\caption{Trans. RPE RMSE (m/s)}
\label{tab2}  
\begin{tabular}{p{2.4cm}p{1cm}p{1cm}p{1cm}p{1cm}}
\hline\noalign{\smallskip}
\textbf{Sequence}  & JF & SF & FF & PF  \\
\noalign{\smallskip}\hline\noalign{\smallskip}
\hline    
      fr1/xyz & 0.021& \textbf{0.012} &0.023 &0.019\\
      fr1/desk2 & 0.084& 0.041 & 0.038& \textbf{0.031} \\
\hline
 fr3/walk\_xyz& 0.68  & 0.29 & 0.21 & \textbf{0.13}\\
     fr3/walking\_static & 0.18 & 0.097 &\textbf{0.030}& 0.072 \\
     HRPSlam2.1 & 0.35 & 0.32 &\textbf{0.28} &0.31\\
     HRPSlam2.4 & \textbf{0.31} & 0.63 &0.59 & 0.41\\
     HRPSlam2.6 & 0.12 & 0.11 &\textbf{0.060} &0.10\\ 
\noalign{\smallskip}\hline\noalign{\smallskip}
\end{tabular}
\vspace{-0.5cm}
\end{table}
To compute the ATE of one trajectory, firstly align it to the ground truth using the least-square method, and then directly compares the distances between the estimated positions and the ground truth at the same timestamps.
The RPE is the relative pose error at timestamp over a time interval.
Our experiments are implemented on a desktop that has Intel Xeon(R) CPU E5-1620 v4 @ 3.50 GHz $\times$ 8, 64 GiB System memory and dual GeForce GTX 1080 Ti GPUs. The experimental setting of FF is given in Tab.\ref{tab:para-list}. We set the image pyramid levels as 4, each level at max 2 iterations, thus, the total iteration times limitation is 8. For the comparison experiments, we adopt their default parameters.

We first evaluate the proposed dynamic segmentation method on TUM RGB-D dynamic sequence fr3/walking\_xyz, which contains 827 RGB-D images, including two moving humans and slightly object motions (\eg the chairs are slightly moved by the people). See Figure \ref{fig:tumseg}, which indicates the dynamic segmentation performances of JF, SF, PF and FF. The first row is input RGB frames, the other rows are the dynamic/static segments of each method, the last column shows the background reconstruction results(except JF, since their open source version didn't provide reconstruction function).
See Tab.\ref{tab1} and Tab.\ref{tab2} for the comparison results. In which, in the static sequences $fr1/xyz$ and $fr1/desk2$, the VO performances of these four methods are similar, because SF, PF and FF are all basing on EF framework. EF is dedicated to static(or slightly dynamic) local areas reconstruction, thus in static sequences, these three methods are all converge to EF's performance.

In the highly dynamic sequences, these four methods show different pros and cons. 
Our previous work PF achieves very small errors in the scenes which only contain human objects. Depending on the deep learning based detection method, PF detect both dynamic and static human objects with clear segment boundaries(see the fourth row in Fig.\ref{fig:tumseg}). However, the drawback is that PF always tends to segments the PCDs attached to the humans into foreground segments, see the wrong segmentation on table and chairs areas close to the humans objects. 
Furthermore, as PF's object detection front-end OpenPose \cite{op} doesn't work well if the input image has no head, PF dropped its performances in HRPSlam sequences(Because in HRPSlam datasets, the camera was mounted on a 151 cm high humanoid robot, who cannot smooth inspect human faces). 
JF and SF detect the moving objects by jointly minimize the intensity and depth energy function, but these energy functions lack of items with dynamic property, which leads to wrong dynamic/static segmentation in the 2nd and 3rd rows of Fig.\ref{fig:tumseg}. 
As the proposed FF involved the optical flow residuals which greatly indicate the pixel's moving status, FF achieved dynamic object extraction in frame 4 and 606 and reduced the wrong static background segmentation as shown in frame 184, 506 and 606. 

Images in Fig.\ref{fig:tumplot} plot the ATE and RPE of the fr3/walking\_xyz sequence. 
In which, PF achieved the smallest trajectory errors. Amongst the module-free dynamic SLAM methods, the proposed FF outperforms the others. 
PF achieved very small trajectory errors, root-mean-square-error (rmse) 4.1 cm, while FF gets 12 cm.

These results indicate that the proposed optical flow residuals based static/dynamic semantic segmentation method achieved efficient dynamic foreground PCDs extraction performances in RGB-D benchmarks. 
Similar to PF, FF performs as similar as EF in the static scenes. 
The advantage of FF is not relying on object modules. 
FF can extract different kinds of moving objects, while PF can only detect human objects. The disadvantage of FF (same to the other model-free methods, \eg SF, JF) is non-sensitive to slight motions, neither very fast motions, such as robot falling down. As shown in Figure \ref{fig:hrpslam}, since very fast motions usually result in wrong optical flow estimations.  


\section{Conclusions}
In this paper, we provided a novel dense RGB-D SLAM algorithm that jointly figures out the dynamic segments and reconstructs the static environments. The newly provided dynamic segmentation and dense fusion formulation applied the advanced dense optical flow estimator, which enhanced the dynamic segmentation performance in both accuracy and efficiency. The demonstrations on both online datasets and real robotics application scenes showed competitive performances in both static and dynamic environments.
\section*{Acknowledgements}
This work was supported by JSPS Grants-in-Aid for Scientific Research (A) 17H06291. 
We thank Dr. Raluca Scona for opening source the codes of StaticFusion. The discussion with her is very helpful.
\bibliographystyle{IEEEtran}
\bibliography{refs}

\begin{thebibliography}{10}
\providecommand{\url}[1]{#1}
\csname url@samestyle\endcsname
\providecommand{\newblock}{\relax}
\providecommand{\bibinfo}[2]{#2}
\providecommand{\BIBentrySTDinterwordspacing}{\spaceskip=0pt\relax}
\providecommand{\BIBentryALTinterwordstretchfactor}{4}
\providecommand{\BIBentryALTinterwordspacing}{\spaceskip=\fontdimen2\font plus
\BIBentryALTinterwordstretchfactor\fontdimen3\font minus
  \fontdimen4\font\relax}
\providecommand{\BIBforeignlanguage}[2]{{%
\expandafter\ifx\csname l@#1\endcsname\relax
\typeout{** WARNING: IEEEtran.bst: No hyphenation pattern has been}%
\typeout{** loaded for the language `#1'. Using the pattern for}%
\typeout{** the default language instead.}%
\else
\language=\csname l@#1\endcsname
\fi
#2}}
\providecommand{\BIBdecl}{\relax}
\BIBdecl

\bibitem{dynamicSLAM-survey}
M.~R.~U. Saputra, A.~Markham, and N.~Trigoni, ``Visual slam and structure from
  motion in dynamic environments: A survey,'' \emph{ACM Computing Surveys
  (CSUR)}, vol.~51, no.~2, p.~37, 2018.

\bibitem{kf}
R.~A. Newcombe, S.~Izadi, O.~Hilliges, D.~Molyneaux, D.~Kim, A.~J. Davison,
  P.~Kohi, J.~Shotton, S.~Hodges, and A.~Fitzgibbon, ``Kinectfusion: Real-time
  dense surface mapping and tracking,'' in \emph{Mixed and augmented reality
  (ISMAR), 2011 10th IEEE international symposium on}.\hskip 1em plus 0.5em
  minus 0.4em\relax IEEE, 2011, pp. 127--136.

\bibitem{ef}
T.~Whelan, R.~F. Salas-Moreno, B.~Glocker, A.~J. Davison, and S.~Leutenegger,
  ``Elasticfusion: Real-time dense slam and light source estimation,''
  \emph{The International Journal of Robotics Research}, vol.~35, no.~14, pp.
  1697--1716, 2016.

\bibitem{cf}
M.~R{\"u}nz and L.~Agapito, ``Co-fusion: Real-time segmentation, tracking and
  fusion of multiple objects,'' \emph{arXiv preprint arXiv:1706.06629}, 2017.

\bibitem{mid-f}
B.~Xu, W.~Li, D.~Tzoumanikas, M.~Bloesch, A.~Davison, and S.~Leutenegger,
  ``Mid-fusion: Octree-based object-level multi-instance dynamic slam,'' in
  \emph{2019 International Conference on Robotics and Automation (ICRA)}.\hskip
  1em plus 0.5em minus 0.4em\relax IEEE, 2019, pp. 5231--5237.

\bibitem{cf-seg}
P.~O. Pinheiro, T.-Y. Lin, R.~Collobert, and P.~Doll{\'a}r, ``Learning to
  refine object segments,'' in \emph{European Conference on Computer
  Vision}.\hskip 1em plus 0.5em minus 0.4em\relax Springer, 2016, pp. 75--91.

\bibitem{jf}
M.~Jaimez, C.~Kerl, J.~Gonzalez-Jimenez, and D.~Cremers, ``Fast odometry and
  scene flow from rgb-d cameras based on geometric clustering,'' in \emph{2017
  IEEE International Conference on Robotics and Automation (ICRA)}.\hskip 1em
  plus 0.5em minus 0.4em\relax IEEE, 2017, pp. 3992--3999.

\bibitem{sf}
R.~Scona, M.~Jaimez, Y.~R. Petillot, M.~Fallon, and D.~Cremers, ``Staticfusion:
  Background reconstruction for dense rgb-d slam in dynamic environments,'' in
  \emph{2018 IEEE International Conference on Robotics and Automation
  (ICRA)}.\hskip 1em plus 0.5em minus 0.4em\relax IEEE, 2018, pp. 1--9.

\bibitem{pf}
T.~Zhang and Y.~Nakamura, ``Posefusion: Dense rgb-d slam in dynamic human
  environments,'' in \emph{Proceedings of the 2018 International Symposium on
  Experimental Robotics}, J.~Xiao, T.~Kr{\"o}ger, and O.~Khatib, Eds.\hskip 1em
  plus 0.5em minus 0.4em\relax Cham: Springer International Publishing, 2020,
  pp. 772--780.

\bibitem{ds}
C.~Yu, Z.~Liu, X.-J. Liu, F.~Xie, Y.~Yang, Q.~Wei, and Q.~Fei, ``Ds-slam: A
  semantic visual slam towards dynamic environments,'' in \emph{2018 IEEE/RSJ
  International Conference on Intelligent Robots and Systems (IROS)}.\hskip 1em
  plus 0.5em minus 0.4em\relax IEEE, 2018, pp. 1168--1174.

\bibitem{segnet}
V.~Badrinarayanan, A.~Kendall, and R.~Cipolla, ``Segnet: A deep convolutional
  encoder-decoder architecture for image segmentation,'' \emph{IEEE
  transactions on pattern analysis and machine intelligence}, vol.~39, no.~12,
  pp. 2481--2495, 2017.

\bibitem{orb2}
R.~Mur-Artal and J.~D. Tard{\'o}s, ``Orb-slam2: An open-source slam system for
  monocular, stereo, and rgb-d cameras,'' \emph{IEEE Transactions on Robotics},
  vol.~33, no.~5, pp. 1255--1262, 2017.

\bibitem{rigidity}
J.~Wulff, L.~Sevilla-Lara, and M.~J. Black, ``Optical flow in mostly rigid
  scenes,'' in \emph{IEEE Conference on Computer Vision and Pattern Recognition
  (CVPR)}, vol.~2, no.~3.\hskip 1em plus 0.5em minus 0.4em\relax IEEE, 2017,
  p.~7.

\bibitem{lvzhaoyang}
Z.~Lv, K.~Kim, A.~Troccoli, D.~Sun, J.~M. Rehg, and J.~Kautz, ``Learning
  rigidity in dynamic scenes with a moving camera for 3d motion field
  estimation,'' \emph{arXiv preprint arXiv:1804.04259}, 2018.

\bibitem{pwc}
D.~Sun, X.~Yang, M.-Y. Liu, and J.~Kautz, ``Pwc-net: Cnns for optical flow
  using pyramid, warping, and cost volume,'' in \emph{Proceedings of the IEEE
  Conference on Computer Vision and Pattern Recognition}, 2018, pp. 8934--8943.

\bibitem{sv}
J.~Papon, A.~Abramov, M.~Schoeler, and F.~Worgotter, ``Voxel cloud connectivity
  segmentation-supervoxels for point clouds,'' in \emph{Computer Vision and
  Pattern Recognition (CVPR), 2013 IEEE Conference on}.\hskip 1em plus 0.5em
  minus 0.4em\relax IEEE, 2013, pp. 2027--2034.

\bibitem{robust}
C.~Kerl, J.~Sturm, and D.~Cremers, ``Robust odometry estimation for rgb-d
  cameras,'' in \emph{Robotics and Automation (ICRA), 2013 IEEE International
  Conference on}.\hskip 1em plus 0.5em minus 0.4em\relax IEEE, 2013, pp.
  3748--3754.

\bibitem{dual}
M.~Jaimez, M.~Souiai, J.~Gonzalez-Jimenez, and D.~Cremers, ``A primal-dual
  framework for real-time dense rgb-d scene flow,'' in \emph{Robotics and
  Automation (ICRA), 2015 IEEE International Conference on}.\hskip 1em plus
  0.5em minus 0.4em\relax IEEE, 2015, pp. 98--104.

\bibitem{tum}
J.~Sturm, N.~Engelhard, F.~Endres, W.~Burgard, and D.~Cremers, ``A benchmark
  for the evaluation of rgb-d slam systems,'' in \emph{Intelligent Robots and
  Systems (IROS), 2012 IEEE/RSJ International Conference on}.\hskip 1em plus
  0.5em minus 0.4em\relax IEEE, 2012, pp. 573--580.

\bibitem{hrpslam}
T.~Zhang and Y.~Nakamura, ``Hrpslam: A benchmark for rgb-d dynamic slam and
  humanoid vision,'' in \emph{2019 Third IEEE International Conference on
  Robotic Computing (IRC)}.\hskip 1em plus 0.5em minus 0.4em\relax IEEE, 2019,
  pp. 110--116.

\bibitem{op}
Z.~Cao, T.~Simon, S.~Wei, and Y.~Sheikh, ``Realtime multi-person 2d pose
  estimation using part affinity fields,'' in \emph{{CVPR}}.\hskip 1em plus
  0.5em minus 0.4em\relax {IEEE} Computer Society, 2017, pp. 1302--1310.

\end{thebibliography}
\end{document}